\documentclass[runningheads]{llncs}
\usepackage{graphicx}
\usepackage{amsmath}
\usepackage{amssymb}
\usepackage{multirow}
\usepackage{booktabs}
\usepackage{pifont}
\usepackage{textcomp}
\usepackage{hyperref}
\newcommand\methodname{APEx}
\begin{document}
\title{Anatomy-guided Pathology Segmentation}
%
%
\author{Alexander Jaus
	\inst{1,2}
	\and
	Constantin Seibold \inst{3,4}
	\and Simon Reiß \inst{1} 
	\and Lukas Heine \inst{3}
	\and Anton Schily \inst{3}
	\and Moon Kim \inst{3}
	\and Fin Hendrik Bahnsen \inst{3}
	\and Ken Herrmann \inst{4,5}
	\and Rainer Stiefelhagen* \inst{1}
	\and Jens Kleesiek* \inst{3}
}

\authorrunning{A. Jaus et al.}

\institute{Karlsruhe Institute of Technology, Karlsruhe, Germany
	\email{\{firstname.lastname\}@kit.edu} 
	\and
	HIDSS4Health - Helmholtz Information and Data Science School for Health, Karlsruhe/Heidelberg, Germany
	\and
	Institute for AI in Medicine, University Hospital Essen, Essen, Germany
	\email{\{firstname.lastname\}@uk-essen.de}
	\and
	Department of Nuclear Medicine, University of Duisburg-Essen
	\email{\{firstname.lastname\}@uk-essen.de}
	\and
	German Cancer Consortium (DKTK)- University Hospital Essen, Essen, Germany
}

\renewcommand{\thefootnote}{\fnsymbol{footnote}}
\footnotetext[1]{shared last author}
\maketitle              
\begin{abstract}
Pathological structures in medical images are typically deviations from the expected anatomy of a patient. 
While clinicians consider this interplay between anatomy and pathology, recent deep learning algorithms specialize in recognizing either one of the two, rarely considering the patient's body from such a joint perspective.
In this paper, we develop a generalist segmentation model that combines anatomical and pathological information, aiming to enhance the segmentation accuracy of pathological features.
Our Anatomy-Pathology Exchange (APEx) training utilizes a query-based segmentation transformer which decodes a joint feature space into query-representations for human anatomy and interleaves them via a mixing strategy into the pathology-decoder for anatomy-informed pathology predictions. 
In doing so, we are able to report the best results across the board on FDG-PET-CT and Chest X-Ray pathology segmentation tasks with a margin of up to $3.3\%$ as compared to strong baseline methods. Code and models will be publicly available at \href{https://github.com/alexanderjaus/APEx}{github.com/alexanderjaus/APEx}.
\end{abstract}
\section{Introduction}
Throughout their extensive training, radiologists acquaint themselves with human biology and physiology, enabling them to discern typical patterns in the anatomy of both healthy individuals and those presenting health concerns.
Years of clinical practice empower doctors to use this underlying knowledge about the body to associate very nuanced visual anatomy abnormalities with specific diseases correctly. 
This holistic approach of doctors, considering both anatomy and pathology in the tissue is contrasted by the vast amount of current automatic pathology segmentation models that specialize in narrow disease types and fall short of an overall understanding of body structures~\cite{schultheiss2021lung,menze2014multimodal}. 
These models are generally end-to-end semantic segmentation learners~\cite{ronneberger2015unet,oktay2018attention}, and resemble models designed for the natural image domain and as such could be applied interchangeably in both domains, from pathology- to street-scene-~\cite{zhang2018road} and everyday object segmentation~\cite{gamal2018shuffleseg,wang2018depth}.
Conversely, the medical imaging field has an obvious, yet often disregarded continuity which is -- of course -- the context is always the human body with the patient's anatomy.

While patients' anatomical features vary, the medical biases that associate anatomy with pathology for radiological assessment remain constant, such as simple observations, that a fracture has to be associated with a bone structure or that tumor locations often correspond to anatomical regions.
When identifying a pathology, current segmentation models might or might not pick up anatomy-pathology correlations during training, which is the reverse direction to using anatomical priors for pathology identification. In the spirit of a doctor's workflow, we ask: Can explicitly learned human anatomy improve a model's capability to predict pathological structures?

Within this work, we explore different strategies to incorporate anatomical knowledge which we model as anatomical labels to improve upon pathology predictions. Inspired by the training of medical professionals, we propose a joint training procedure in which our network learns to predict both: anatomy and pathology via our proposed \methodname~ architecture.



We summarize our contributions as follows:
(1) We ablate multiple strategies on how to incorporate learned anatomical knowledge into pathology segmentation models. 
(2) We introduce a query-based, joint anatomy and pathology instance segmentation model \methodname, which is capable of enriching pathology predictions by integrating anatomy knowledge via shared embeddings and query mixing.
(3) We validate the performance of \methodname\ on two different medical datasets covering whole-body FDG-PET-CT and chest X-Ray with as diverse anatomical structures as bones, organs and vessels for an improved joint anatomy-pathology recognition of $+2.0\%$, $+3.3\%$ respectively. 
\subsubsection{Related Work:}
Leveraging anatomical knowledge as a prior has been addressed in some previous works in the form of shapes~\cite{yao2019integrating,oktay2017anatomically,navarro2019shape}, expected textures~\cite{ibragimov2017combining} or atlas-based segmentation models~\cite{huang2021medical} to aid segmentation. While these works mostly aim to segment anatomical structures, some works have started to leverage anatomical priors to improve upon pathology detection in X-Ray~\cite{muller2023anatomy} or Colorectal Cancer segmentation in CT~\cite{zhang2023ag}. These studies highlight the beneficial effects of using anatomical priors but mainly utilize manually designed features tailored towards a specific use-case (e.g. the intestinal wall is important to detect colorectal cancer). 

Within this work, instead of following the idea of manually designed features, 
we opt for a more data-driven and deep-learning-inspired approach: We investigate the usefulness of learned anatomical features to aid the segmentation of pathology. This approach breaks free of hand-crafted prior limitations and allows us to capture the knowledge available thanks to recently available holistic anatomical datasets~\cite{wasserthal2023totalsegmentator,jaus2023towards,seibold2023accurate}. We hypothesize that utilizing anatomical features helps identify pathologies as deviations from the expected anatomy.




\section{Methodology}
\noindent In this section, we first present the learning setup for anatomy and pathology segmentation and walk through our ablations to incorporate anatomical knowledge into the model training. Finally, we derive our so called \textbf{A}natomy-\textbf{P}athology \textbf{Ex}change (\textbf{\methodname})~strategy to jointly learn both anatomy and pathology.

\subsection{Preliminaries}

Our formulation of the anatomy and pathology segmentation task depends on a training dataset:

\begin{equation}
    \mathcal{D} = \{ (x_i, a_i, p_i) \}_{i=0}^N \enspace,
\end{equation}
with $x_i \in \mathbb{R}^{3 \times H \times W}$ referring to one of the $N$ images in the dataset, while $a_i \in \lbrack 0, \dots, A\rbrack^{H \times W}$ is the associated anatomy with $A$ classes and $p_i \in \lbrack 0, \dots, P\rbrack^{H \times W}$ the pathology mask with $P$ classes within the image.
The task of a trained model is to predict, for new unseen test images $x_t$ for each pixel in the image the correct anatomy categories $a_t$ as well as the correct pathology classes $p_t$. 

If the dataset provides instance-level annotations, we extend the approach to an instance-aware regime. Each anatomical mask $a_i$ and pathological mask $p_i$ then includes not only class- but instance-aware targets.

To investigate whether anatomical knowledge aids in identifying deviations from expected anatomy, we will examine two different tasks in two distinct domains: semantic segmentation of cancer in PET-CT images and instance-aware segmentation of thoracic abnormalities in chest X-Rays.

To accommodate these varied requirements, we opt for a 2D model due to the constraints of the X-Ray domain and model the 3D PET-CT images as sliced 2D images. To address the differing demands of semantic and instance-aware segmentation, we align with recent advancements in segmentation literature~\cite{cheng2021mask2former,li2023mask,carion2020end,zhu2020deformable} which intertwine both semantic- and instance segmentation through the design choice of predicting high-dimensional query vectors, which combined with pixel-wise embeddings, encode instance-wise segments in an image.
These queries are then employed to classify each segment, encapsulating information about both the segment's class and its shape. As a starting point for the experiments, we choose a Mask2Former~\cite{cheng2021mask2former} architecture. Our chosen setup is flexible in the choice of image modalities and in the choice of segmentation tasks. 

\subsection{Incorporating Learned Anatomical Knowlege: A roadmap}
To investigate how to incorporate anatomical knowledge into the model training, we perform several ablations in a five-fold cross-validation setting in the domain of PET-CT. The baseline comparison model is a Mask2Former~\cite{cheng2021mask2former} model trained only on pathological labels. We report the 5-fold Validation IoU scores of naive anatomy incooperation techniques in Tab.~\ref{tab:ablations} (left).

\begin{table}
\caption{Val scores on the 5-fold CV PET-CT splits. A. Cond, A. Pred and $\gamma$ denote anatomy conditioning, auxillary anatomy learning and a weight factor respectively.}

\centering
    \begin{tabular}{lcccc}
   \toprule
   \multicolumn{5}{c}{\textbf{Naive Anatomy Incooperation}}\\
   Method & A. Cond & A. Pred &$\gamma$ & IoU\\
   \midrule
   Baseline & -- & -- & -- & $\phantom{0}54.34 \pm 1.46$\\
   Pretrain & \checkmark & -- & -- & $\phantom{0}56.64 \pm 3.06$\\
   Multitask & -- & \checkmark & 1 &  $\phantom{0}56.10 \pm 3.36$ \\
   Multitask & -- & \checkmark & 10 &  $\phantom{0}57.12 \pm 4.17$ \\
   Multitask & -- & \checkmark & 142 &  $\phantom{0}55.89 \pm 3.03$ \\
   Ana In & \checkmark & -- & -- &  $\phantom{0}57.23 \pm 2.71$ \\
   Ana In & \checkmark & \checkmark & 1 &  $\phantom{0}56.52 \pm 4.14$ \\
   \bottomrule
    \end{tabular}
    \begin{tabular}{lc}
    \toprule
    \multicolumn{2}{c}{\textbf{Architecture Ablations}}\\
    Method & IoU \\
    \midrule
         Baseline & $ 54.34 \pm 1.46$\\
         +Shared BB & $ 54.44 \pm 4.14$\\
         +Shared PD &  $ 58.69 \pm 3.63$\\
         \quad $^\llcorner$Query Sum & $ 59.56 \pm 3.64$ \\
         \quad $^\llcorner$Query Sum 2-ways & $ 59.35 \pm 3.18$ \\
         \quad $^\llcorner$Query Mean & $ 59.78 \pm 3.23$ \\
          \quad $^\llcorner$Cross Attention (CA) & $ 59.42 \pm 2.42$\\
          \quad $^\llcorner$CA per feature level & $ 58.48 \pm 2.52$ \\
         \bottomrule
    \end{tabular}
    \label{tab:ablations}
\end{table}

\noindent First, we investigate the effect of pretraining on anatomy. This leads to an improvement of about $2.3\%$.
\newline
\textbf{Multitask Prediction:} Next, we compare to jointly learned features using a multi-task setting approach. We paste the pathological labels atop the anatomical labels predicting an additional class. Despite being suboptimal, since PET-CT pixels could be interpreted as both, anatomy and pathology, depending on the context, this leads to a similar improvement as pretraining. However, treating pathology as just another class underestimates its significance. To address this, we apply a weighted loss with weight $\gamma$, amplifying the pathology class's importance by 10-fold and 142-fold to equate it with the 142 anatomical labels. The 10-fold increase yields positive results, whereas the 142-fold adjustment demonstrates the challenge of selecting an appropriate weight factor. \newline
\textbf{Anatomy as an Auxiliary Input:} Inspired by atlas-based segmentation methods, we input anatomical labels along with the PET-CT image mimicking an optimal anatomy atlas. Using this procedure, we receive similar results as with the previous approach. 
\subsubsection{Architecture Ablations:}
The last section's analysis underscores that while anatomical knowledge can enhance pathology prediction, its effective utilization is complex. 
Thus, in our second experimental series, we postulate that due to the inherent overlap between anatomical and pathological labels, a two-head prediction approach is optimal. \newline
Initially, only the ResNet50 backbone is shared between the two prediction heads, resulting in no major improvement. A critical adjustment involves the sharing of a PixelDecoder across both anatomical and pathological prediction tasks. This integration significantly boosts the performance, evidenced by a notable increase of over $4\%$ in IoU. This enhancement underscores the PixelDecoder's role in generating pixel embeddings rich in anatomical and pathological information, marking it as a crucial element in our design reflecting the dual role of each pixel in this task. \newline
\textbf{Query Mixing Strategies:}
Ultimately, as we employ distinct transformer decoders for anatomical and pathological predictions, we probe the efficacy of information exchange mechanisms via query exchange. This reflects the possibility of a direct exchange of queries representing anatomical and pathological segments. We explore various strategies, including nonparametric mixing and more flexible communication strategies such as cross-attention. While almost all strategies lead to a positive effect, none of them shines out as a clear winner. 

\begin{figure*}[t]
    \centering
    \includegraphics[width=0.95\linewidth, 
    ]{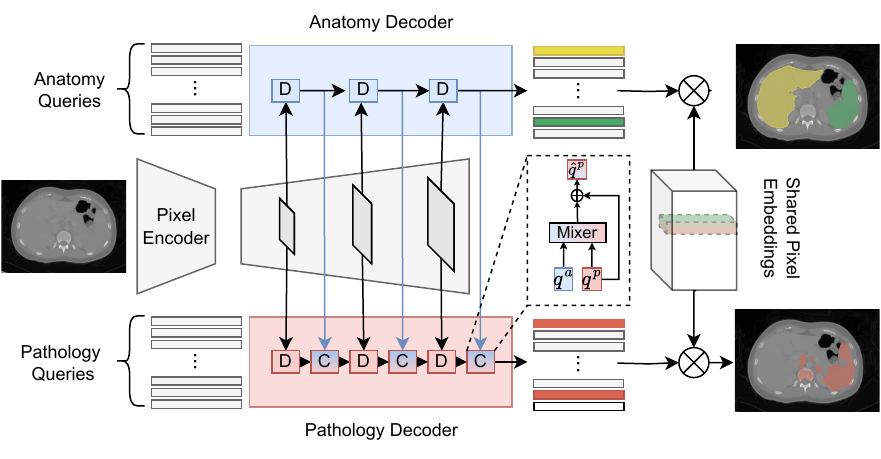}
    \caption{Overview of the proposed \methodname\ Method, leveraging a shared pixel encoder, shared pixel embedding space, separate decoders and a query-mixing module.
    }
    \label{fig:method}
\end{figure*}


We conclude this section with the insight that a two-head prediction one for the anatomy and one for pathologies leveraging shared pixel-embedding is a crucial design choice. On top, enabling communication between the different decoders leads to a further performance boost. The best-ablated model performs about $5.44\%$ better than the naive baseline model.


\subsection{Proposed Approach: \methodname}

\noindent   \methodname~is based on a query-based segmentation approach leveraging anatomical and pathological information. It incorporates anatomical context via the exchange of information between two decoders: One tasked to segment the anatomy and one tasked to segment the pathology. We show the overall method in Fig.~\ref{fig:method}.

\subsubsection{Shared Embedding Architecture:}

\noindent Starting with a standard $2$D image $x\in \mathbb{R}^{3\times H \times W}$ we encode the image using a feature extractor $f^{extr}$ (parameterized by a ResNet50~\cite{he2016deep}), which maps $x$ to a set of feature maps at different scales $f^{extr}(x) = \{ F_i\}_{i=0}^n$ with $F_i \in \mathbb{R}^{H_i \times W_i}$, such that $H_i > H_{i + 1}$ and $W_i > W_{i+1}$ hold, i.e., feature maps successively get smaller in spatial extent.
These feature maps are then decoded using an arbitrary pixel-decoder. We choose to use the deformable DETR~\cite{zhu2020deformable} model as a pixel-decoder producing a set of enriched pixel embeddings $\{J_i\}_{i=0}^n$, with $J_i \in \mathbb{R}^{d \times H_i \times W_i}$.

\subsubsection{Anatomy and Pathology Decoders:}

Our architecture is motivated in computing separate query vectors for anatomy and pathology classes and let the anatomy queries influence the pathology queries while limiting the reverse influence only to a shared embedding space.

Each enriched pixel encoding map $J_i$ is accessed by two decoding functions $f^{ana}_i(\cdot)$ and $f^{path}_i(\cdot)$ from the function sets $\{f^{ana}_i(\cdot)\}_{i=n}^1$ and $\{f^{path}_i(\cdot)\}_{i=n}^1$ which either decode the anatomy or the pathology from it.



Randomly initialized, but learnable parameter-queries $q_0^{ana}$ and $q_0^{path}$ are transformed via 
\begin{align}
    q_{i+1}^{ana} &= f^{ana}_i(q_{i}^{ana}, J_{i}) \enspace\text{and} \\ \enspace
    q_{i+1}^{path} &= f^{path}_i(q_{i}^{path}, J_{i})\enspace,
\end{align}
and optimized during training. 
The decoders $f_i(\cdot)$ follow a standard masked transformer setup,~i.e. queries are transformed through a cross-attention layer that attends to the joint embeddings of the respective scale $i$, followed by a self-attention and feed-forward layer.
For the pathology branch $\{f^{path}_i(\cdot)\}_{i=n}^1$ to explicitly adhere to the learned anatomical queries an anatomy-to-pathology communication strategy is designed next.


\subsubsection{Anatomy to Pathology Communication Strategy}

Medical personnel have access to a large amount of knowledge regarding the human body, which current pathology segmentation models do not have. Besides the implicit information exchange via the shared pixel embedding, we propose to integrate a communication step $f^{mix}_i(\cdot)$ after each pathology-decoder step $f_i^{path}(\cdot)$.
There the queries $q_i^{path}$ resulting from the scale $i$ pathology-decoder are enriched with the anatomy queries $q_i^{ana}$ from the anatomy-decoder as follows:
\begin{align}
    \label{Eq: Merge_queries}
    \hat{q}^{path}_i = f_i^{mix}(q^{ana}_i, q^{path}_i) 
\end{align}
Here, $\hat{q}^{path}_i$ is the anatomy-enriched pathology query which, through a mixing strategy is capable of capturing anatomical information. We did not find a superior mixing strategy and thus would either recommend averaging the queries as a nonparametric approach or using a cross-attention mixing module.

In this asymmetric architectural setup, anatomical information influences the pathology-specific queries while the anatomy branch stays agnostic to any pathology and simply reflects the patient-specific anatomy details serving as a useful foundational prior in pathology assessment. This design is ablated against an inferior design in which the anatomy branch is updated by the pathology as well (cf. Tab.~\ref{tab:ablations}: Query Sum 2-ways).  






\subsubsection{Joint Anatomy and Pathology Segmentation}
Bringing the whole architecture and processing steps together into our Anatomy and Pathology Exchange (\methodname) pipeline, we predict the anatomy and pathology segments through the following dot product:
\begin{align}
    out^{ana} &= J_0 \cdot q^{ana}_{n-1} \enspace\text{and} \\ \enspace
    out^{path} &= J_0 \cdot \hat{q}^{path}_{n-1} \enspace,
\end{align}
Query vectors are passed through a simple classifier to associate anatomy or pathology classes to the predicted segments. 
The parameters of all components, namely $f^{extr}$, $f^{ana}$, $f^{path}$, and $f^{mix}$ are optimized via weighted cross-entropy and binary mask losses enforced on each anatomy and pathology prediction $out^{ana}$ and $out^{path}$.







\section{Experiments and Results}
\begin{table}[t]
\renewcommand{\arraystretch}{0.9}
\centering
\caption{Comparison of \methodname\ against multiple SOTA methods in the PET-CT domain (left). We highlight the \textbf{best} and the \underline{second best} performance.}
\label{tab: autopet oct results}
\small
\begin{tabular}{lccccccc}
\toprule
\multirow{2}{*}{Method} & \multicolumn{2}{c}{\textbf{PET-CT VAL}} & & \multicolumn{2}{c}{\textbf{PET-CT TEST}} \\
& IoU & BIoU && IoU & BIoU \\
\midrule
DLV3+\cite{chen2018encoder} & $55.00 \pm 3.5$ & $54.78 \pm 3.6$  && $53.60 \pm 5.4$ & $53.07 \pm 5.4$  \\
M2F\cite{cheng2021mask2former} & $54.34 \pm 1.4$ & $54.16 \pm 1.6$  && $55.48 \pm 1.1$ & $55.02 \pm 1.1$  \\
UNET    \cite{ronneberger2015unet} & $\underline{57.62 \pm 3.2}$ & $\underline{57.38 \pm 3.3}$  && $\underline{56.43 \pm 1.5}$ & $\underline{55.86 \pm 1.4}$ \\
\midrule
Ours (CA) & \textbf{$\mathbf{59.43 \pm 2.43}$} & \textbf{$\mathbf{59.21 \pm 2.62}$}  && \textbf{$\mathbf{57.5 \pm 0.88}$} & \textbf{$\mathbf{57.04 \pm 0.89}$} \\
\bottomrule
\end{tabular}

\end{table}

\begin{figure} [b]
    
    \begin{tabular}{ccc|cc}
  \hline
  \textbf{Target} & \textbf{M2F~\cite{cheng2021mask2former}} &\textbf{\methodname\ (ours)}  & 
   \multicolumn{2}{c}{\textbf{~\methodname\ Attended Anatomy} }\\
    \midrule
    \includegraphics[width=0.18\linewidth, trim=20 340 655 0, clip]{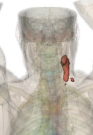} &
    \includegraphics[width=0.18\linewidth, trim=20 340 655 0, clip]{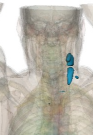} &
    \includegraphics[width=0.18\linewidth, trim=20 340 655 0, clip]{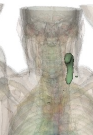} & 
        \includegraphics[width=0.18\linewidth, trim=20 310 635 0, clip]{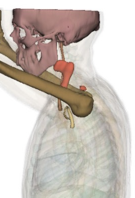} &
    \includegraphics[width=0.18\linewidth, trim=20 310 635 0, clip]{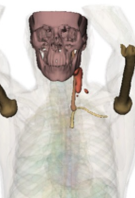}

    \end{tabular}

  \caption{Stacked 2D tumor predictions next to top-5 attended anatomical structures.}
  \label{fig: qual_autopet_attention}
\end{figure}

\noindent\textbf{Datasets}:
To assess our method's broad applicability, we performed experiments across two vastly different medical imaging domains: FDG-PET-CT, and Chest X-Ray. For FDG-PET-CT, due to the absence of a comprehensive dataset with both anatomical and lesion annotations, we merged two distinct datasets: autoPET~\cite{gatidis2022whole}, which provides lesion annotations, and the Atlas dataset~\cite{jaus2023towards}, offering anatomical details. We exclude patients without pathologies, motivated by the high accuracy ($\geq 95\%$) of binary classifiers for cancer detection in PET images. Our study utilized $185$ 3D volumes for five-fold cross-validation and an additional test set of $125$ cancer patients from the remaining dataset. To adapt images to the selected 2D setting, we slice them axially and stack CT and PET images channel-wise, leaving the third channel empty.

In the X-Ray domain, we evaluate the properties of our method on the ChestXDet\cite{liu2020chestxdet10} dataset containing 13 pathology classes. To train anatomy segmentation, we predict anatomy pseudo-labels onto this dataset using a model trained on the PaxRay++ dataset~\cite{seibold2023accurate}. We evaluate the different methods using five-fold cross-validation on the training set. During training, we omit images with no pathologies.






\noindent\textbf{Baselines and Methods}:
When evaluating models across different domains, we determine the best performing candidate based on the performance on the individual validation sets. We use either the official test splits, if they exist, or a test set that we reserved beforehand. We benchmark \methodname\  on PET-CT against established $2$D segmentation baselines such as UNet \cite{ronneberger2015unet}, DeeplabV3+ \cite{chen2018encoder} and Mask2Former \cite{cheng2021mask2former}. In all experiments, we ensure models are trained using identical data and learning pipelines to isolate the effect of incorporating anatomical knowledge. For chest X-Ray, we compare on instance segmentation against PointRend~\cite{kirillov2019pointrend}, MaskDino~\cite{li2023mask} and Mask2Former\cite{cheng2021mask2former}. Regarding the specific \methodname\ architecture, we choose the Cross-Attention Query Mixer, as it offered a competitive performance with the lowest standard deviation during our initial ablations (cf. Tab.~\ref{tab:ablations}).


\subsection{Semantic- and Instance Segmentation Results}
\noindent\textbf{PET-CT Results:}
In Tab.~\ref{tab: autopet oct results} we report the IoU and Boundary IoU~\cite{cheng2021boundary} (BIoU) performances of the previously mentioned baseline segmentation models against our method. All models have been initialized with LVM-MED weights~\cite{nguyen2023lvm} to provide a fair comparison. 
The results indicate that our method is capable of outperforming multiple strong competitors on our five-fold validation splits and the holdout testset. Fig.~\ref{fig: qual_autopet_attention} shows qualitative results as well as the most attended anatomical structures during the cross-attention query mixing step.

\noindent\textbf{ChestXDet Results:} In Tab.~\ref{tab:chestx_result}, we show the performance of different state-of-the-art instance segmentation methods trained using the same backbone. We see that our method improves over the Mask2Former-baseline by $\sim$3.75\% mAP. Across 12 of 13 pathologies, our method achieves the best, or second-best performance, improving over recent transformer architectures as well as established CNN models. Detailed results and qualitative examples are in the Appendix.

\begin{table*}[t]
\label{tab:chestx_result}
\centering
\scriptsize
\caption{ChestXDet\cite{liu2020chestxdet10} results. We highlight the best performance in \textbf{bold} and the second best by \underline{underlining}. Detailed results in Appendix. }
\begin{tabular}{lccccccc}
\toprule
Pathology & MRCNN\cite{he2017mask} &CascCRCNN~\cite{cai2018cascade}&PointRend\cite{kirillov2019pointrend} & MskDino\cite{li2023mask} & M2F\cite{cheng2021mask2former} & Ours (CA)  \\
\midrule
mAP (Val) & $13.98 \pm 0.40$& $14.64\pm0.44$ & $15.33\pm 0.82$ & $16.44 \pm	1.43$ & \underline{$16.57 \pm	0.67$}  & $\mathbf{17.16 \pm 0.63}$ \\

mAP (Test) & $13.72 \pm 0.41$& $13.86\pm0.70$ & \underline{$15.14\pm 0.44$} & $14.38 \pm	0.74$ & $13.87 \pm 0.53$ & $\mathbf{17.20 \pm 0.33}$ \\
\bottomrule
\end{tabular}

\end{table*}

\section{Conclusion}
We proposed a novel way of leveraging anatomical information to improve pathology segmentation and showed the efficacy of the general concept of anatomy-guidance in two different domains covering diverse anatomical structures and pathologies.
Besides improved performance, our method \methodname~encourages the exchange of anatomical information to ensure pathology segments are informed by the patient's anatomy, aligning more with the workflow of doctors that developed over decades.

\bibliographystyle{splncs04}
%
\bibliography{bibliography}

\end{document}